\documentclass[letterpaper, 10 pt, conference]{ieeeconf}
\IEEEoverridecommandlockouts
\usepackage{cite}
\usepackage{amsmath,amssymb,amsfonts}
\usepackage{algorithm}
\usepackage{algorithmic}
\usepackage{graphicx}
\usepackage{textcomp}
\usepackage{tabularx}

\usepackage{pifont}
\newcommand{\cmark}{\textcolor{Green}{\ding{51}}}%
\newcommand{\xmark}{\textcolor{Red}{\ding{55}}}%
\usepackage[dvipsnames]{xcolor}
\usepackage{diagbox}
\usepackage{adjustbox}
\usepackage{multirow}
\usepackage[utf8]{inputenc}
\usepackage{booktabs, makecell}
\usepackage{subfigure}
\usepackage{hyperref}
\let\labelindent\relax
\usepackage[inline]{enumitem}

\usepackage{threeparttable}
\usepackage{blindtext}
\usepackage{geometry}
\def\BibTeX{{\rm B\kern-.05em{\sc i\kern-.025em b}\kern-.08em
    T\kern-.1667em\lower.7ex\hbox{E}\kern-.125emX}}
\begin{document}

\title{\LARGE \textbf{A Survey of Federated Learning for Connected and Automated Vehicles}}
\author{
Vishnu Pandi Chellapandi, Liangqi Yuan, Stanislaw H \.{Z}ak and Ziran Wang
\thanks{V. P. Chellapandi, L. Yuan, S. H. \.{Z}ak and Z. Wang are with College of Engineering, Purdue University, West Lafayette, IN 47907, USA. Emails: {\tt\small \{cvp,yuan383,zak,ziran\}@purdue.edu}}
}

\maketitle
\begin{abstract}
Connected and Automated Vehicles (CAVs) are one of the emerging technologies in the automotive domain that has the potential to alleviate the issues of accidents, traffic congestion, and pollutant emissions, leading to a safe, efficient, and sustainable transportation system. Machine learning-based methods are widely used in CAVs for crucial tasks like perception, motion planning, and motion control, where machine learning models in CAVs are solely trained using the local vehicle data, and the performance is not certain when exposed to new environments or unseen conditions. Federated learning (FL) is an effective solution for CAVs that enables a collaborative model development with multiple vehicles in a distributed learning framework. FL enables CAVs to learn from a wide range of driving environments and improve their overall performance while ensuring the privacy and security of local vehicle data. In this paper, we review the progress accomplished by researchers in applying FL to CAVs. A broader view of the various data modalities and algorithms that have been implemented on CAVs is provided. Specific applications of FL are reviewed in detail, and an analysis of the challenges and future scope of research are presented.
\end{abstract}


\newcommand{\etal}{et al.}

\section{Introduction}
\label{Sec. Introduction}
Connected and automated vehicles (CAVs) are the key to future intelligent transport systems. With the advent of big data, the Internet of things (IoT), edge computing, and intelligent systems CAVs have the potential to improve the efficiency of the overall transportation system, and reduce traffic congestion and accidents. Robust network communication and significantly increased internet speed are expected to be guaranteed with the onset of advanced communication infrastructures. Currently, CAVs are generating a tremendous amount of raw data, up to one to two terabytes per vehicle per day \cite{sterk_monetizing_2022} from various sources like engine components, electronic control units (ECU), perception sensors, and vehicle-to-everything (V2X) communications. This large amount of data is sent to the cloud continuously or periodically for monitoring, prognostics, diagnostics, and connectivity features. These data are also private and come under strict privacy protection regulations in various regions. One such example is the General Data Protection Regulation (GDPR) in the European Union \cite{voigt2017eu}. Even with the development of advanced machine learning (ML) techniques and vehicle connectivity, it has not been feasible to have a centralized framework to collect data from every vehicle and train an ML model securely. These limitations led to the development of a new ML paradigm known as Federated Learning (FL) \cite{kairouz_advances_2021,yang_federated_2019}. 


FL is a new technology breakthrough that has been actively implemented in several application domains. FL has been coined by Google~\cite{mcmahan_communication-efficient_2017} and was initially used for mobile keyboard prediction in Gboard~\cite{hard_federated_2019} to allow multiple mobile phones to cooperatively and securely train a neural network (NN) model. In FL, the edge devices/clients only send the gradients or the learnable parameters to the cloud server rather than sending massive local datasets in a Centralized Learning (CL) framework. The cloud server performs a secure aggregation \cite{bonawitz2017practical} of the received gradients/weights and updates the global model parameters that are transmitted back to the clients/edge devices. This procedure, known as a communication round, continues iteratively until the convergence criteria are met in the global model optimization. The key advantage of FL is reducing the strain on the network while also preserving the privacy of the local data. FL is a potential candidate that can utilize the data available from each CAV and develop a robust ML model. A detailed comparison of the on-device vehicle training, CL, and FL approach is described in Table~\ref{Table_Comparison_of_ML_Approaches}.

\begin{table*}[t]
\caption{Comparison of ML Approaches in Connected and Automated Vehicles}
\label{Table_Comparison_of_ML_Approaches}
\centering
\begin{tabularx}{\linewidth}{|l|X|X|X|}
\hline
\textbf{Features} & \textbf{Edge Learning (On-Vehicle only)} & \textbf{Centralized Learning} & \textbf{Federated Learning} \\
\hline
Model training & Local vehicle & Central server & Local vehicle training and central server aggregation \\
\hline
Model applicability & Personalized model & Single global model & Single global model but can be personalized \\
\hline
Privacy protection & \cmark\cmark & \xmark\xmark & \cmark \\
\hline
Learning efficiency & \xmark & \cmark & \cmark\cmark \\
\hline
Performance on heterogeneous/anomaly data & \xmark & \cmark\cmark & \cmark   \\
\hline
Communication (Data transmission) requirement & \cmark\cmark & \xmark\xmark & \xmark \\
\hline
Training data volume & \xmark\xmark & \cmark\cmark & \cmark \\
\hline
Current research progress & \cmark\cmark & \cmark\cmark & \xmark \\
\hline
Compatibility with CAVs & \cmark & \xmark\xmark & \cmark\cmark \\
\hline
\end{tabularx}
\noindent{\cmark\cmark {} best, \cmark {} high, \xmark {} low, \xmark\xmark {} worst.}
\end{table*}

Despite the mutual benefits of connectivity between vehicles, the issues of invasion of privacy, accuracy, effectiveness, and communication resources are essential problems to be addressed. FL frameworks have received attentions for their natural ability to preserve privacy by transmitting only model data between the server and its clients without including user data. In particular, the model data packets are smaller than the user data, thus saving the consumption of communication resources. Similarly, FL frameworks distribute training to each client, and the server does not perform training but only aggregates, which can reduce the pressure on the server and improves training efficiency.

In this survey, we provide an insightful and comprehensive survey of FL for CAV (FL4CAV), including 
diverse applications and 
key challenges. After a review of data modality, data security and FL algorithms in CAV, this survey focus on most of the critical applications of FL4CAV, such as steering wheel angle prediction, vehicle trajectory prediction, object detection, motion control application, and driver monitoring. This survey also highlights the current challenges and future research directions of FL4CAV, such as performance, safety, fairness, applicability, etc.   

The remainder of this paper is organized as follows: Section~\ref{Sec. Overview} highlights the diverse data modalities, data securities, and algorithms of FL in CAVs. Section~\ref{Sec. Applications of FL for CAV} reviews the application of FL in CAVs with detailed examples. The multi-modal data, algorithms, and datasets used in the relevant literature are also summarized. Current challenges and future research directions are discussed in Section~\ref{Sec. Challenges}. Section~\ref{Sec. Conclusion} presents the conclusion of this study and outlines future work. 

\section{Overview of Data Modalities, Data Securities and Algorithms}
\label{Sec. Overview}

\begin{figure}[t]
\centering
\centerline{\includegraphics[width=1\linewidth]{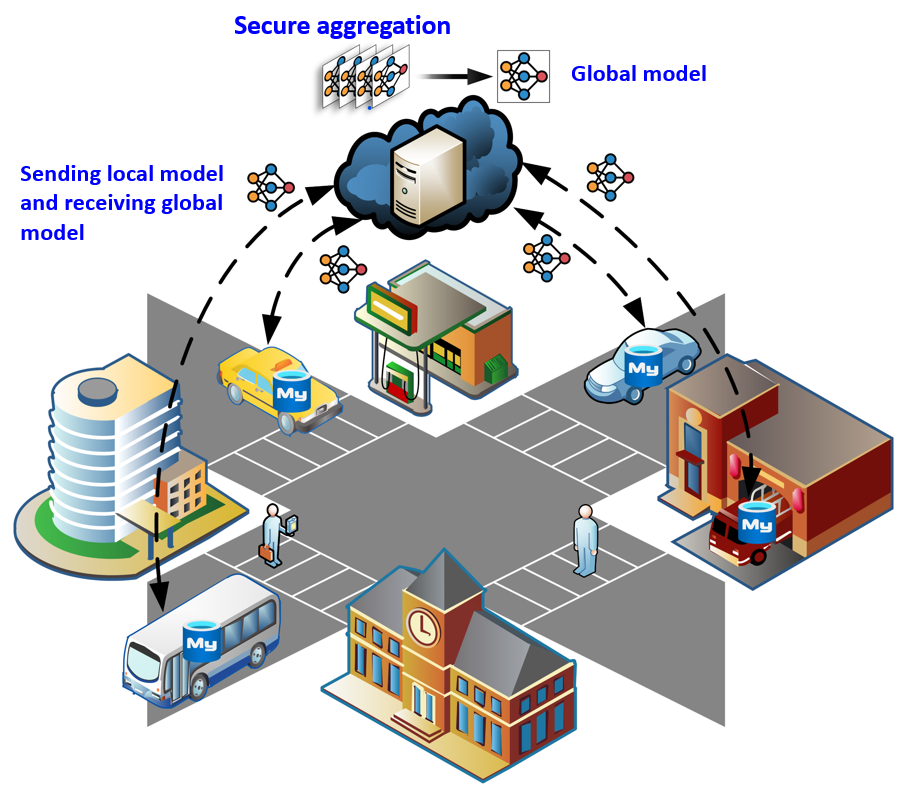}}
\caption{Illustration of FL4CAV.}
\label{Fig. FL_CAV}
\end{figure}

The FL4CAV is illustrated in Fig.~\ref{Fig. FL_CAV}. Each CAV as a client, undertakes sensing data acquisition, signal processing, storage, communication, perception, and decision-making. For sensing data acquisition, a variety of sensors are integrated into CAVs, including global navigation satellite systems (GNSS), multi-modal cameras, Radio Detection And Ranging (RADAR), Light Detection And Ranging (LiDAR), and Inertial Measurement Unit (IMU) to capture the vehicle, driver, passenger, and external information.

The tasks for CAV are also diversified, including target speed tracking, behavior prediction, object detection, driver monitoring, and more. After training on an ML model with local data, clients send the trained model to the server. Then, the server shares a generalized model with clients for perception, prediction, and decision-making purposes. The FL4CAV framework shows a trend towards multi-modal sensing data, massively parallel clients, and multi-class tasks. Recent efforts have been ongoing to understand the applicability and challenges of implementing FL to CAVs \cite{savazzi_opportunities_2021, du_federated_2020, chen_federated_nodate,eskandarian_research_2021, grigorescu_survey_2020}. A detailed review of the data modalities of CAVs, Data security, and FL algorithm is presented below. 


\subsection{Data Modality}

CAVs collect multi-modal data from various sensors for tasks, such as navigation, perception, obstacle detection, and vehicle control. The FL training process involves vehicles that may have a different variety of sensors. The data collected by sensors depends on the sensor type, the sensor's range, the accuracy/precision of the sensor, sensor placement, and the operating environment. The operating environment, such as snow, heavy rain, or fog, can reduce the sensor visibility, thereby deteriorating the data quality. These factors lead to variations that can significantly affect the sensor performance. The performance of the FL model is directly dependent on the quality of the data collected by the vehicles. The data resolution, size, sampling rate, etc., obtained from the CAVs are generally heterogeneous, and processing the data is also a challenging task. Hence a detailed review is presented to understand the various data modalities in FL4CAV applications.

Images, especially visible RGB images, are one of the most important data modalities for CAV. Vision-related tasks such as steering wheel angle prediction~\ref{sec-Steering_wheel_appl}, Traffic sign recognition~\cite{stergiou_federated_2022}, semantic segmentation~\cite{fantauzzo_feddrive_2022}, object detection~\ref{Sec-obj_detection}, and driver monitoring~\cite{yuan2023federated}
typically use images captured by the camera as the data source. In most applications, a variant of the Convolutional Neural Network (CNN) model is trained to achieve the intended functionality. However, due to its intrusive design, privacy issues are always criticized for image-based systems, especially for in-cabin, and driver-related applications \cite{mishra_cabin_2022,doshi2022federated,yuan2023federated,zhao2023fedsup}. FL focuses on the model parameters and ignores the data, which addresses the drawback that visual image-based systems tend to compromise user privacy. Moreover, FL also solves the data transmission problem caused by the inflated data size of images and videos, leading to a more efficient learning framework.

LiDAR data provides a solid foundation for autonomous driving capabilities. LiDAR data have also been utilized for object detection tasks~\cite{rjoub_improving_2021,elbir2022hybrid}. LiDAR generates 3D point clouds that can detect objects accurately even under adverse weather conditions, unlike cameras that are not robust. However, the dense point cloud of LiDAR data makes transmission a daunting task. Therefore, compared with the image-based FL systems, the system of FL for LiDAR data is more interested in improving learning efficiency and saving communication resources.

Vehicle status data is also an important part of the CAV data modality, which reflects more about the vehicle rather than external information. Application of FL such as collision avoidance~\cite{fu_selective_2022}, vehicle trajectory prediction~\ref{sec-veh_traj_appl}, and Motion control application~\ref{sec-motion_control_appl} use vehicle status data like the vehicle coordinates, velocity, acceleration, throttle, and braking. These are typically time-series data at uniform time intervals (200 millisecond/1 second) and can generally be easily processed. Recurrent neural networks (RNN) such as Long Short-term Memory (LSTM) have been used for these applications because of their capability to capture long-term dependencies in the data. Reinforcement learning (RL) is a relatively newer method and is starting to be widely used for various applications such as collision avoidance due to its ability to learn from trial-and-error interactions with the environment, allowing it to adapt to changes in the environment and respond to unseen situations. Vehicle state data is not as strongly motivated as images and LiDAR, but FL can still inspire knowledge sharing among CAVs. For example, sharing contextual knowledge of rare events such as traffic accidents can reduce the risk of CAVs re-encountering the same event.







\subsection{Data Security }
 
Robust and secure privacy-preserving techniques are essential for  protecting sensitive data during the FL process for CAVs. It is demonstrated that the FL process can still be vulnerable to various malicious attacks, such as when one or more participants are compromised, and they could transmit false parameters to hinder the global model performance. The FL central server is also prone to attack and thereby causing the entire learning process to collapse~\cite{bagdasaryan_how_2020,fang_local_2020}.
In this paper, we highlight a few existing solutions, and a detailed review of the techniques is discussed in~\cite{kaissis_secure_2020}. Homomorphic Encryption (HE) is a technique that enables the server to perform the training on encrypted data from the vehicles without the need for decryption. Secure Multi-Party Computation (SMPC) enables multiple vehicles to collaboratively perform computation  with the individual vehicle data without sharing the data with each other. Differential Privacy (DP) is a popular method that ensures privacy in data by adding random noise to the data before being sent to the server to prevent extraction of the information. Blockchain technology-based method is a disruptive technology that has recently been widely deployed in CAV applications~\cite{singh_blockchain_2017,rathee_blockchain_2019,fu_vehicular_2020,pokhrel_federated_2020,basha_inter-locking_2021,he_bift_2022,javed_integration_2022}. Blockchain is a type of digital ledger technology that securely transfer data in a decentralized framework. Data from AVs share their data with the vehicular network, and the information is stored on the blockchain. The system is designed to protect data privacy and data security as well as to provide higher security to the overall vehicular networks engaged in the learning process~\cite{dorri_blockchain_2017}. 

\subsection{Federated Learning Algorithm}

Most of the existing literature uses the FedAvg algorithm~\cite{mcmahan_communication-efficient_2017} for the FL aggregation process in the server---see Table~\ref{Table Related literature}. FedAvg applies Stochastic Gradient Descent (SGD) optimization on local vehicles and does a weighted averaging of the weights from the vehicles in the central server. FedAvg performs multiple local gradient updates before sending the parameters to the server, reducing the number of communication rounds.  For FL4CAV, the data on each CAVs is dynamically updated at each communication round---see Algorithm~\ref{alg:FedAvg}.

\begin{algorithm}[h]
\small
    \caption{FedAvg for Dynamic Data Updating CAVs}
    \label{alg:FedAvg}
    \begin{algorithmic}[1]
       \STATE {\bfseries Input:} Vehicle set $\mathbb{V}$, communication rounds $T$, isolated time-varying local dataset $\xi = \{\xi_v^{(t)} | v \in \mathbb{V}\}$, local epochs $E$, learning rate $\eta$, loss function $f$ \\
       \STATE {\bfseries Output:} Aggregated global model $w$
       \STATE initialize $w_0$ \\
       \FOR{$t = 0, \dots, T-1$} 
       \STATE {\bfseries Perform} local SGD {\bfseries for} vehicle $v \in \mathbb{V} $ {\bfseries in parallel do}
            \STATE \ \ \ Sample $\xi_v^{(t)} \mbox{, compute } g_v^{(t)} := \widetilde{\nabla} f_v (w_v^{(t)},\xi_v^{(t)})$
            \STATE \ \ \ $ w_v^{(t+1)} \ \gets \  w_v^{(t)} - \eta_t g_v^{(t)} \hspace{0.5cm} \implies $ SGD ($E$ epochs)
       \STATE {\bfseries end for}
       \STATE $ w^{(t+1)} \ \gets \ \sum_{v \in \mathbb{V} } \frac{|\xi_v^{(t)}|}{|\xi^{(t)}|} ( w_v^{t+1} ) \hspace{0.12cm} \implies $ FedAvg
       \ENDFOR
    \STATE Output the aggregated global model $w \gets w^{(T)}$
    \end{algorithmic}
\end{algorithm}

Data heterogeneity, client drift, and data imbalance from clients have proved to significantly impact the performance of FedAvg optimization resulting in unstable convergence. The data obtained from CAVs typically have non-independent and identically distributed (non-iid) and imbalanced data distribution. There is a need to develop an FL framework that could perform well with the varying data distribution from CAVs. FedProx \cite{li_federated_2020} algorithm combines FedAvg with a proximal term to improve convergence and reduce communication cost. Fed-ADAM~\cite{reddi_adaptive_2022} has shown improved convergence and optimization performance by incorporating ADAM optimization in FedAvg algorithm. Dynamic Federated Proximal (DFP) algorithm is an extension of the Federated Proximal Algorithm (FPA) that can effectively deal with non-iid data distribution by dynamically varying the learning rate and regularization coefficient during the learning process~\cite{zeng_federated_2022}. Federated Distillation (FD)~\cite{jeong_communication-efficient_2018} uses knowledge distillation to transfer knowledge in a decentralized manner leading to a significant reduction in the communication size compared to a traditional FL and also can have the ability to handle non-iid data samples \cite{liu_communication-efficient_2022}. There have been efforts to address the client heterogeneity, and it is an ongoing research area~\cite{acar_federated_2022,li_convergence_2020,wang_tackling_nodate,wang_field_2021}.


\begin{table*}[t]
\caption{Literature Overview of FL Application to Autonomous Vehicles}
\label{Table Related literature}
\centering
\begin{tabularx}{\linewidth}{|l|l|p{70pt}|p{85pt}|p{60pt}|p{60pt}|X|}
\hline
\textbf{Literature} & \textbf{Time} & \textbf{Data Modality} & \textbf{Application} & \textbf{Base Model} & \textbf{FL Algorithm} & \textbf{Dataset} \\

\hline
\cite{doomra_turn_2020} & 2020 &  Time series data of multiple features from sensors  & Turn signal prediction & LSTM & FedAvg & Ford’s Big Data Drive \cite{doomra_turn_2020} \\
\hline
\cite{zhang_end--end_2021, zhang_real-time_2021} & 2021 & RGB image & Steering angle prediction & Two-stream CNN & Async FL & Self-collected \\
\hline
\cite{m_p_steering_2021} & 2021 & RGB image & Steering angle prediction & Self-defined CNN &  FedAvg & Self-collected\\
\hline
\cite{rjoub_improving_2021} & 2021 & RGB image and Li-
DAR & Object detection & YOLO CNN &  FedSGD & Canadian Adverse Driving Conditions Dataset \cite{pitropov_canadian_2021}\\
\hline
\cite{zeng_federated_2022} & 2022 & RGB image and trajectory data  & Target speed tracking & Self-defined NN & DFP (FedAvg for aggregation) & Berkeley deep drive \cite{yu2020bdd100k} and dataset of annotated car trajectories \cite{moosavi2017trajectory} \\
\hline
\cite{stergiou_federated_2022} & 2022 & RGB image & Traffic sign recognition & LeNet-5 & FedAvg &German Traffic Sign Recognition Benchmark \cite{stallkamp2011german} \\
\hline
\cite{fantauzzo_feddrive_2022} & 2022 &  Multi-modal image  & Semantic Segmentation  & BiSeNet V2 & FedAvg + Variants & Cityscapes \cite{cordts_cityscapes_2016} and IDDA \cite{alberti_idda_2020-1}  \\
\hline
\cite{elbir2022hybrid} & 2022 & RGB image and LiDAR & 3D object detection & U-Net & HFCL (FedAvg for aggregation) & Lyft Level 5 dataset \cite{kesten2019lyft} \\
\hline
\cite{han_federated_2022} & 2022 & Vehicle position, velocity and acceleration + Driver behavior & Trajectory prediction & LSTM & FedAvg+Variants & US‐101 and I‐80 data sets of NGSIM \cite{dot2018next} \\
\hline
\cite{fu_selective_2022} & 2022 & Vehicle position, velocity and acceleration & Collision avoidance & Deep RL & SFRL (FedAvg for aggregation)  & Self-generated \\
\hline
\cite{doshi2022federated} & 2022 & RGB image & Driver activity recognition & ResNet-56 & FedGKT & State Farm Distracted Driver Detection \cite{farm_2016} and AI City Challenge 2022 \cite{Naphade22AIC22} \\
\hline
\cite{yuan2023federated} & 2023 & RGB image & Driver activity recognition & ResNet-34 & FedProx + Variants & State Farm Distracted Driver Detection \cite{farm_2016} and Drive\&Act \cite{martin2019drive} \\
\hline
\cite{zhao2023fedsup} & 2023 & RGB image & Driver fatigue detection & Bayesian CNN & FedSup & Blinking Video Database \cite{pan2007eyeblink} and Eyeblink8 \cite{drutarovsky2014eye} \\
\hline
\end{tabularx}
\end{table*}

\section{Applications of FL for CAV} 
\label{Sec. Applications of FL for CAV}

This section reviews a few important applications in detail of FL in CAVs. The FL4CAV literature, including data modalities, underlying models, applications, and datasets, are highlighted in Table~\ref{Table Related literature}. Different applications on CAV are highly dependent on different strengths of FL, such as protecting privacy, improving learning efficiency, enhancing generalization ability, reducing communication overhead, etc.

\subsection{In-vehicle Human Monitoring} 
\label{Sec. Driver Monitoring}
FL has the potential to enhance the security of user data on board, while enabling knowledge transfer and ensuring the generalization ability of the model. However, in human-related applications where data is highly heterogeneous and personalized, it can be challenging to balance the generalization ability of the model with the need for personalization to specific users.

Driver monitoring applications, such as distraction detection, are critical safety features that monitor the driver's steadiness, and alertness and warn the distracted driver to apply the brakes~\cite{doshi2022federated,zhao2023fedsup}. Driver privacy may be a bigger concern than steering wheel angle prediction and object recognition, leading to FL's ability to be more highlighted in terms of privacy protection. However, the driver monitoring application is a highly personalized application where the driver's behavior is strongly associated with personal habits, emotions, cultural background, and even the interpretation of instructions. This user heterogeneity poses a challenge for FL systems. For human-related applications like driver monitoring, personalized FL is the dominant solution \cite{yuan2023federated}. 

Passenger monitoring applications are an emerging research area that involves detecting passenger intents of boarding and alighting and warning of dangerous behavior in public transportation~\cite{liu2021automatic}. However, this field has not yet received much attention due to the lack of available datasets and the difficulty of monitoring multiple users simultaneously. Nevertheless, the ability of FL to integrate knowledge about public transportation and the growing demand for passenger monitoring makes it a promising application in this area.


\subsection{Steering Wheel Angle Prediction} 
\label{sec-Steering_wheel_appl}
While non-human related applications are generally less invasive to user privacy, they still incur significant learning costs and communication overhead when CAVs are involved. FL remains a strong candidate for these applications, especially when more CAVs are involved.

Steering wheel angle prediction improves driving safety by detecting vehicle yaw as an essential component of Advanced Driver Assistance Systems (ADAS). As of 2020 in the United States, about 30\% of the roads are still unpaved \cite{Highway_Statistics_2020}. Recent technologies like the end-to-end learning approach enable AVs to drive safely even under challenging circumstances, such as unpaved and unmarked roads. The prediction of steering wheel angle is one of the critical aspects of an autonomous vehicle (Society of Automotive Engineers (SAE) Level 4 and Level 5 \cite{J3016_201806}) that controls the lateral position of the vehicle. The steering wheel angle is predicted from the RGB images collected from the front-facing camera as input. The dataset consisting of camera images and the steering wheel angle is trained on a CNN model \cite{gidado_survey_2020}. The CNN models have been demonstrated to efficiently perform a lane following task on an unmarked road \cite{bojarski2016end}. Related literature has demonstrated that FL can collaboratively learn the neural network model at a significantly lower communication cost while also preserving privacy~\cite{zhang_end--end_2021, zhang_real-time_2021,m_p_steering_2021}. 

\subsection{Vehicle Trajectory Prediction} 
\label{sec-veh_traj_appl}
FL can improve the model's ability to handle rare events, such as traffic accidents, adverse weather conditions, and risky behaviors, by leveraging the collective knowledge of CAVs. 

Vehicle trajectory prediction allows drivers or ADAS to anticipate potentially dangerous behavior of other vehicles, such as sudden lane changes, skidding, or tire blowouts, in order to react proactively and prevent accidents. Autonomous vehicles navigate in highly-uncertain and interactive environments shared with other dynamic agents. In order to plan safe and comfortable maneuvers, they need to predict the future trajectories of surrounding vehicles. The inherent uncertainty of the future makes trajectory prediction a challenging task~\cite{huang_survey_2022}. AVs are often required to drive in a dynamic and challenging environment along with other vehicles. In these scenarios, predicting the trajectories of the surrounding vehicles/environment are crucial for safer and more comfortable navigation. Accurate prediction of trajectories is one of the complex tasks of an AV due to high computation cost, diverse driving styles (aggressive/defensive), dynamic behavior of the road obstacles (vehicles, pedestrians, objects), and several noise factors (sensors, weather). The task of predicting an optimal vehicle trajectory is challenging, but the use of a collaborative learning platform can facilitate the development of an efficient model for safe driving. By enabling models to learn from rare hazard events of each CAVS, they can react in a timely manner to avoid potential accidents and create a safer driving environment. In~\cite{han_federated_2022, rjoub_explainable_2022}, it has been reported that the FL approach achieves a similar performance over centralized learning.



\subsection{Object Detection} 
\label{Sec-obj_detection}
FL enables the CAV framework to learn efficiently with low communication overhead, which is particularly advantageous when the volume of data is much larger than the size of the model.


Object detection is one of the most important functions of the visual perception system, and FL can effectively help CAVs detect diverse objects in different driving scenarios, such as road, traffic, vehicles, obstacles, and pedestrians. Most of the current techniques use 2D detection and 3D detection methods. The 3D detection methods provide additional information, such as the sizes, locations, and classes of the objects that are necessary for motion planning, collision avoidance, and motion control. 3D object detection has witnessed significant advances due to the rapid evolution of deep learning-based methods~\cite{arnold_survey_2019}. However, the data size of 3D object detection, such as LiDAR and high resolution image, is typically large. As a result, there are significant challenges to deploying robust object detection models on a traditional centralized learning approach due to the high communication and computation overhead. These concerns can be mitigated through the use of a FL based approach for CAVs. 

FL has already been in practice much before for computer vision-related tasks such as developing safety hazard warning solutions in smart city applications \cite{liu_fedvision_2020}. Object detection accuracy generally struggles under adverse weather circumstances such as snow. It has demonstrated that the CNN-FL framework improves the detection accuracy and performs better than the centralized and gossip decentralized models~\cite{rjoub_improving_2021}. Recently there have been numerous studies to improve the performance of FL on complex tasks like object detection~\cite{wang2022edge}. A hybrid federated and centralized learning (HFCL) framework was proposed that allows vehicles with computational resources to be part of the FL training process while the others transmit their local data to the server like a centralized learning process. The trade-off between the computational and communication overhead of the vehicles is addressed. The performance of HFCL is not shown to be better than a centralized learning approach in this example \cite{elbir2022hybrid} and is a subject for further research and improvement. It is demonstrated that with multi-stage resource allocation and robust device selection, the performance of FL significantly improved compared to traditional centralized learning and baseline FL approaches~\cite{wang_federated_2022}.



\subsection{Motion Control Application} 
\label{sec-motion_control_appl}
FL enables CAVs to quickly adapt to different driving scenarios, including unfamiliar and unvisited roads, cities, and countries. Additionally, FL may enable CAVs to adjust driving styles based on different driving habits, climate, scenarios, and cultural norms.


Motion control application, including learning-based methods such as RL, and Genetic Algorithm (GA) have proved effective in solving difficult and challenging control-related tasks~\cite{perez-gil_deep_2022}. The controls typically include throttle, steering wheel angle, and braking. It has been demonstrated that reinforcement methods-based driving achieves better performance than human-based driving. However, implementing RL-based methods on vehicles is challenging due to the high computational power demand. 

FL approach enables CAVs to train and optimize controller parameters collaboratively. A few potential benefits are enabling AVs to adapt to unseen routes/traffic scenarios or operating conditions because of past data from other AVs, on-ramp acceleration, driving in congested traffic scenarios, and so on. For instance, if a driver from a rural area is driving in a city, FL can help the CAV quickly adapt to the new driving environment and enhance the safety, comfort, energy efficiency, and effectiveness of the driving experience. Target speed tracking is one application that has used FL recently to perform target speed tracking under various scenarios. In~\cite{zeng_federated_2022}, a two-layer Multi-layer Perceptron (MLP) model was trained in an FL framework to dynamically adjust the Proportional Integral Derivative (PID) gains controller parameters to achieve the desired target speed efficiently under various traffic scenarios.



\section{Challenges} 
\label{Sec. Challenges}
In this section, we review and summarize the remaining challenges in state-of-the-art technology and the future scope of research.

\subsection{Resource Limitations}
\textbf{Massively parallel CAVs raise questions about collaboration capabilities, management, and resources.} Huge CAVs participation in the FL could increase the solve time, memory utilization, and therefore the computational power for the global model update. In particular, the vision-related perception tasks have concerns such as high communication costs and not being flexible towards heterogeneous datasets. Decentralized FL and Clustered FL~\cite{taik_clustered_2022} are also being explored to reduce the communication overhead.

\textbf{Lack of sufficient real-world datasets, simulator, and pre-trained base model.} There is a need for more real-world datasets (different weather conditions and traffic scenarios), a realistic high-fidelity FL4CAV simulator for seamless FL integration~\cite{wang_federated_2022}, and a good pre-training model. Federated transfer learning is a new approach that has been adopted to improve the model performance, and accuracy \cite{basha_inter-locking_2021,yuan2023federated}.

\subsection{Imperfect Methodology}
 \textbf{Privacy and security issues.} Massive data also leads to privacy and security concerns. This problem must be addressed to train the ML model efficiently without compromising on the model's accuracy and redundancy.
  
\textbf{Lack of robust approach for vehicle selection and resource allocation.} Currently, there is no popular mechanism that can select non-redundant data from CAVs to minimize the network strain. There are ongoing efforts to develop robust methods to select vehicles and resource allocation schemes~\cite{wang_content-based_2021,tianqing2021resource,albelaihi_green_2022}. In~\cite{nishio_client_2019}, the overall training process was demonstrated to be efficient due to incorporating a client selection model. The setup looks at the resource availability of the clients and then determines the clients eligible to be part of the FL global model learning process.  In~\cite{rjoub_explainable_2022}, it is demonstrated that the model performance was improved with AVs that were selected by a trust-based deep reinforcement. 

\textbf{Catastrophic forgetting.} CAV cannot keep all user data due to storage capacity limitations, and new data will always be generated during iteration. Therefore, when the FL framework is updated on new data in iteration---see Algorithm \ref{alg:FedAvg}, the global model forgets the previous knowledge and leads to catastrophic forgetting.

\textbf{Lack of robust fairness and incentive mechanism.} Need for a robust rewarding mechanism. The amount of information shared by CAVs is different and highly inconsistent (Data imbalance). Hence, there needs to be a fair incentive mechanism to reward CAVs for their contribution.

\subsection{Inadequate Evaluation Criteria}
\textbf{FL suitability evaluation for new users.} It is often difficult for the newcomer vehicle to make any informed decisions. In~\cite{rjoub_explainable_2022}, a trust-aware Deep RL model is proposed to assist new vehicles in making superior trajectory and motion planning decisions. 
  
\textbf{Need for high capability diagnostics.} There are several noise factors that could influence the decision of the FL, such as faulty sensors in a visual perception case and incorrect imputation of missing data. The development of a robust diagnostic that can identify and eliminate the updates from these vehicles is needed.

\section{Conclusion}
\label{Sec. Conclusion}
In summary, FL is a new technology breakthrough and has started to be applied in the CAV domain. This paper reviews the various developments, data modalities, and algorithms of FL4CAV and provides a broad list of applications of FL in CAVs. In particular, our focus was on current challenges and the future scope of FL4CAV.

Future work lies in continuing the effort in this paper to provide a more detailed, in-depth, and comprehensive survey of FL4CAV and related vehicle fields. Especially some of the most advanced technologies aim to address communication, security and privacy, and system heterogeneity issues. Some feasible solutions, such as fully decentralized FL and model compression, are feasible for communication problems. Current techniques of threats and malicious attacks on FL are also open problems, such as backdoor attacks, free-riding attacks, and eavesdropping. Some protective techniques, such as differential privacy techniques and homomorphic encryption, are used as potential solutions for security and privacy. Our focus is to apply vehicular FL by drawing on relevant research in other areas, such as healthcare, IoT, and industry. At the same time, a matching FL framework needs to be developed, taking into account the communication, computation, storage, and usage scenarios underlying the vehicles.

\bibliographystyle{ieeetr}
\bibliography{ITSC.bib}

\begin{thebibliography}{10}

\bibitem{sterk_monetizing_2022}
F.~Sterk, D.~Dann, and C.~Weinhardt, ``Monetizing {{Car Data}}: {{A Literature
  Review}} on {{Data-Driven Business Models}} in the {{Connected Car
  Domain}},'' in {\em Hawaii {{International Conference}} on {{System
  Sciences}}}, 2022.

\bibitem{voigt2017eu}
P.~Voigt and A.~Von~dem Bussche, ``The eu general data protection regulation
  (gdpr),'' {\em A Practical Guide, 1st Ed., Cham: Springer International
  Publishing}, vol.~10, no.~3152676, pp.~10--5555, 2017.

\bibitem{kairouz_advances_2021}
P.~Kairouz, H.~B. McMahan, {\em et~al.}, ``Advances and open problems in
  federated learning,'' {\em Foundations and Trends® in Machine Learning},
  vol.~14, no.~1–2, pp.~1--210, 2021.

\bibitem{yang_federated_2019}
Q.~Yang, Y.~Liu, T.~Chen, and Y.~Tong, ``Federated machine learning:
  {{Concept}} and applications,'' {\em ACM Transactions on Intelligent Systems
  and Technology (TIST)}, vol.~10, no.~2, pp.~1--19, 2019.

\bibitem{mcmahan_communication-efficient_2017}
B.~McMahan, E.~Moore, D.~Ramage, S.~Hampson, and B.~A. y~Arcas,
  ``Communication-{{Efficient Learning}} of {{Deep Networks}} from
  {{Decentralized Data}},'' in {\em Proceedings of the 20th {{International
  Conference}} on {{Artificial Intelligence}} and {{Statistics}}},
  pp.~1273--1282, {PMLR}, Apr. 2017.

\bibitem{hard_federated_2019}
A.~Hard, K.~Rao, R.~Mathews, F.~Beaufays, S.~Augenstein, H.~Eichner, C.~Kiddon,
  and D.~Ramage, ``Federated learning for mobile keyboard prediction,'' {\em
  CoRR}, vol.~abs/1811.03604, 2018.

\bibitem{bonawitz2017practical}
K.~Bonawitz, V.~Ivanov, B.~Kreuter, A.~Marcedone, H.~B. McMahan, S.~Patel,
  D.~Ramage, A.~Segal, and K.~Seth, ``Practical secure aggregation for
  privacy-preserving machine learning,'' in {\em proceedings of the 2017 ACM
  SIGSAC Conference on Computer and Communications Security}, pp.~1175--1191,
  2017.

\bibitem{savazzi_opportunities_2021}
S.~Savazzi, M.~Nicoli, M.~Bennis, S.~Kianoush, and L.~Barbieri, ``Opportunities
  of {{Federated Learning}} in {{Connected}}, {{Cooperative}}, and {{Automated
  Industrial Systems}},'' {\em IEEE Communications Magazine}, vol.~59,
  pp.~16--21, Feb. 2021.

\bibitem{du_federated_2020}
Z.~Du, C.~Wu, T.~Yoshinaga, K.-L.~A. Yau, Y.~Ji, and J.~Li, ``Federated
  {{Learning}} for {{Vehicular Internet}} of {{Things}}: {{Recent Advances}}
  and {{Open Issues}},'' {\em IEEE Open Journal of the Computer Society},
  vol.~1, pp.~45--61, 2020.

\bibitem{chen_federated_nodate}
L.~Chen, M.~Torstensson, and C.~Englund, ``Federated learning to enable
  automotive collaborative ecosystem: Opportunities and challenges,'' {\em
  Virtual ITS European Congress}, Nov. 2020.

\bibitem{eskandarian_research_2021}
A.~Eskandarian, C.~Wu, and C.~Sun, ``Research {{Advances}} and {{Challenges}}
  of {{Autonomous}} and {{Connected Ground Vehicles}},'' {\em IEEE Transactions
  on Intelligent Transportation Systems}, vol.~22, pp.~683--711, Feb. 2021.

\bibitem{grigorescu_survey_2020}
S.~Grigorescu, B.~Trasnea, T.~Cocias, and G.~Macesanu, ``A {{Survey}} of {{Deep
  Learning Techniques}} for {{Autonomous Driving}},'' {\em Journal of Field
  Robotics}, vol.~37, pp.~362--386, Apr. 2020.

\bibitem{stergiou_federated_2022}
K.~D. Stergiou, K.~E. Psannis, V.~Vitsas, and Y.~Ishibashi, ``A {{Federated
  Learning Approach}} for {{Enhancing Autonomous Vehicles Image
  Recognition}},'' in {\em 2022 4th {{International Conference}} on {{Computer
  Communication}} and the {{Internet}} ({{ICCCI}})}, ({Chiba, Japan}),
  pp.~87--90, {IEEE}, July 2022.

\bibitem{fantauzzo_feddrive_2022}
L.~Fantauzzo, E.~Fan{\`\i}, D.~Caldarola, A.~Tavera, F.~Cermelli, M.~Ciccone,
  and B.~Caputo, ``{FedDrive: Generalizing federated learning to semantic
  segmentation in autonomous driving},'' in {\em 2022 IEEE/RSJ International
  Conference on Intelligent Robots and Systems (IROS)}, pp.~11504--11511, IEEE,
  2022.

\bibitem{yuan2023federated}
L.~Yuan, L.~Su, and Z.~Wang, ``Federated transfer-ordered-personalized learning
  for driver monitoring application,'' {\em arXiv preprint arXiv:2301.04829},
  2023.

\bibitem{mishra_cabin_2022}
A.~Mishra, S.~Lee, D.~Kim, and S.~Kim, ``In-cabin monitoring system for
  autonomous vehicles,'' {\em Sensors}, vol.~22, no.~12, 2022.

\bibitem{doshi2022federated}
K.~Doshi and Y.~Yilmaz, ``Federated learning-based driver activity recognition
  for edge devices,'' in {\em Proceedings of the IEEE/CVF Conference on
  Computer Vision and Pattern Recognition}, pp.~3338--3346, 2022.

\bibitem{zhao2023fedsup}
C.~Zhao, Z.~Gao, Q.~Wang, K.~Xiao, Z.~Mo, and M.~J. Deen, ``Fedsup: A
  communication-efficient federated learning fatigue driving behaviors
  supervision approach,'' {\em Future Gener. Comput. Syst.}, vol.~138,
  pp.~52--60, Jan. 2023.

\bibitem{rjoub_improving_2021}
G.~Rjoub, O.~A. Wahab, J.~Bentahar, and A.~S. Bataineh, ``{Improving Autonomous
  Vehicles Safety in Snow Weather using Federated YOLO CNN Learning},'' in {\em
  Mobile Web and Intelligent Information Systems: 17th International
  Conference, MobiWIS 2021, Virtual Event, August 23--25, 2021, Proceedings},
  pp.~121--134, Springer, 2021.

\bibitem{elbir2022hybrid}
A.~M. Elbir, S.~Coleri, A.~K. Papazafeiropoulos, P.~Kourtessis, and
  S.~Chatzinotas, ``A hybrid architecture for federated and centralized
  learning,'' {\em IEEE Trans. Cogn. Commun. Netw.}, Jun. 2022.

\bibitem{fu_selective_2022}
Y.~Fu, C.~Li, F.~R. Yu, T.~H. Luan, and Y.~Zhang, ``A {{Selective Federated
  Reinforcement Learning Strategy}} for {{Autonomous Driving}},'' {\em IEEE
  Transactions on Intelligent Transportation Systems}, pp.~1--14, 2022.

\bibitem{bagdasaryan_how_2020}
E.~Bagdasaryan, A.~Veit, Y.~Hua, D.~Estrin, and V.~Shmatikov, ``How to backdoor
  federated learning,'' in {\em International {{Conference}} on {{Artificial
  Intelligence}} and {{Statistics}}}, pp.~2938--2948, {PMLR}, 2020.

\bibitem{fang_local_2020}
M.~Fang, X.~Cao, J.~Jia, and N.~Gong, ``Local model poisoning attacks to
  {{Byzantine-Robust}} federated learning,'' in {\em 29th {{USENIX Security
  Symposium}} ({{USENIX Security}} 20)}, pp.~1605--1622, 2020.

\bibitem{kaissis_secure_2020}
G.~A. Kaissis, M.~R. Makowski, D.~R{\"u}ckert, and R.~F. Braren, ``{Secure,
  Privacy-Preserving and Federated Machine Learning in Medical Imaging},'' {\em
  Nature Machine Intelligence}, vol.~2, pp.~305--311, June 2020.

\bibitem{singh_blockchain_2017}
M.~Singh and S.~Kim, ``Blockchain based intelligent vehicle data sharing
  framework,'' {\em arXiv preprint arXiv:1708.09721}, 2017.

\bibitem{rathee_blockchain_2019}
G.~Rathee, A.~Sharma, R.~Iqbal, M.~Aloqaily, N.~Jaglan, and R.~Kumar, ``A
  blockchain framework for securing connected and autonomous vehicles,'' {\em
  Sensors}, vol.~19, no.~14, p.~3165, 2019.

\bibitem{fu_vehicular_2020}
Y.~Fu, F.~R. Yu, C.~Li, T.~H. Luan, and Y.~Zhang, ``Vehicular
  {{Blockchain-Based Collective Learning}} for {{Connected}} and {{Autonomous
  Vehicles}},'' {\em IEEE Wireless Communications}, vol.~27, pp.~197--203, Apr.
  2020.

\bibitem{pokhrel_federated_2020}
S.~R. Pokhrel and J.~Choi, ``Federated {{Learning With Blockchain}} for
  {{Autonomous Vehicles}}: {{Analysis}} and {{Design Challenges}},'' {\em IEEE
  Transactions on Communications}, vol.~68, pp.~4734--4746, Aug. 2020.

\bibitem{basha_inter-locking_2021}
S.~M. Basha, S.~T. Ahmed, N.~S.~N. Iyengar, and R.~D. Caytiles,
  ``Inter-{{Locking Dependency Evaluation Schema}} based on {{Block-chain
  Enabled Federated Transfer Learning}} for {{Autonomous Vehicular Systems}},''
  in {\em 2021 {{Second International Conference}} on {{Innovative Technology
  Convergence}} ({{CITC}})}, ({Sibalom, Philippines}), pp.~46--51, {IEEE}, Dec.
  2021.

\bibitem{he_bift_2022}
Y.~He, K.~Huang, G.~Zhang, F.~R. Yu, J.~Chen, and J.~Li, ``Bift: {{A
  Blockchain-Based Federated Learning System}} for {{Connected}} and
  {{Autonomous Vehicles}},'' {\em IEEE Internet of Things Journal}, vol.~9,
  pp.~12311--12322, July 2022.

\bibitem{javed_integration_2022}
A.~R. Javed, M.~A. Hassan, F.~Shahzad, W.~Ahmed, S.~Singh, T.~Baker, and T.~R.
  Gadekallu, ``Integration of blockchain technology and federated learning in
  vehicular (iot) networks: {{A}} comprehensive survey,'' {\em Sensors},
  vol.~22, no.~12, p.~4394, 2022.

\bibitem{dorri_blockchain_2017}
A.~Dorri, M.~Steger, S.~S. Kanhere, and R.~Jurdak, ``{{BlockChain}}: {{A
  Distributed Solution}} to {{Automotive Security}} and {{Privacy}},'' {\em
  IEEE Communications Magazine}, vol.~55, pp.~119--125, Dec. 2017.

\bibitem{li_federated_2020}
T.~Li, A.~K. Sahu, M.~Zaheer, M.~Sanjabi, A.~Talwalkar, and V.~Smith,
  ``Federated {{Optimization}} in {{Heterogeneous Networks}},'' {\em
  Proceedings of Machine Learning and Systems}, vol.~2, pp.~429--450, Mar.
  2020.

\bibitem{reddi_adaptive_2022}
S.~J. Reddi, Z.~Charles, M.~Zaheer, Z.~Garrett, K.~Rush, J.~Kone{\v c}n{\'y},
  S.~Kumar, and H.~B. McMahan, ``Adaptive {{Federated Optimization}},'' in {\em
  International {{Conference}} on {{Learning Representations}}}, Mar. 2022.

\bibitem{zeng_federated_2022}
T.~Zeng, O.~Semiari, M.~Chen, W.~Saad, and M.~Bennis, ``{Federated Learning on
  the Road Autonomous Controller Design for Connected and Autonomous
  Vehicles},'' {\em IEEE Transactions on Wireless Communications}, vol.~21,
  no.~12, pp.~10407--10423, 2022.

\bibitem{jeong_communication-efficient_2018}
E.~Jeong, S.~Oh, H.~Kim, J.~Park, M.~Bennis, and S.-L. Kim,
  ``Communication-efficient on-device machine learning: Federated distillation
  and augmentation under non-iid private data,'' {\em ArXiv},
  vol.~abs/1811.11479, 2018.

\bibitem{liu_communication-efficient_2022}
L.~Liu, J.~Zhang, S.~H. Song, and K.~B. Letaief, ``Communication-efficient
  federated distillation with active data sampling,'' in {\em ICC 2022 - IEEE
  International Conference on Communications}, pp.~201--206, 2022.

\bibitem{acar_federated_2022}
D.~A.~E. Acar, Y.~Zhao, R.~Matas, M.~Mattina, P.~Whatmough, and V.~Saligrama,
  ``Federated {{Learning Based}} on {{Dynamic Regularization}},'' in {\em
  International {{Conference}} on {{Learning Representations}}}, Feb. 2022.

\bibitem{li_convergence_2020}
X.~Li, K.~Huang, W.~Yang, S.~Wang, and Z.~Zhang, ``On the {{Convergence}} of
  {{FedAvg}} on {{Non-IID Data}},'' in {\em International {{Conference}} on
  {{Learning Representations}}}, Mar. 2020.

\bibitem{wang_tackling_nodate}
J.~Wang, Q.~Liu, H.~Liang, G.~Joshi, and H.~V. Poor, ``Tackling the objective
  inconsistency problem in heterogeneous federated optimization,'' {\em
  Advances in Neural Information Processing Systems}, vol.~33, pp.~7611--7623,
  2020.

\bibitem{wang_field_2021}
J.~Wang, Z.~Charles, {\em et~al.}, ``A field guide to federated optimization,''
  {\em CoRR}, vol.~abs/2107.06917, 2021.

\bibitem{doomra_turn_2020}
S.~Doomra, N.~Kohli, and S.~Athavale, ``{Turn Signal Prediction: {A} Federated
  Learning Case Study},'' {\em CoRR}, vol.~abs/2012.12401, 2020.

\bibitem{zhang_end--end_2021}
H.~Zhang, J.~Bosch, and H.~H. Olsson, ``End-to-{{End Federated Learning}} for
  {{Autonomous Driving Vehicles}},'' in {\em 2021 {{International Joint
  Conference}} on {{Neural Networks}} ({{IJCNN}})}, ({Shenzhen, China}),
  pp.~1--8, {IEEE}, July 2021.

\bibitem{zhang_real-time_2021}
H.~Zhang, J.~Bosch, and H.~H. Olsson, ``Real-time {{End-to-End Federated
  Learning}}: {{An Automotive Case Study}},'' in {\em 2021 {{IEEE}} 45th
  {{Annual Computers}}, {{Software}}, and {{Applications Conference}}
  ({{COMPSAC}})}, ({Madrid, Spain}), pp.~459--468, {IEEE}, July 2021.

\bibitem{m_p_steering_2021}
M.~Aparna, R.~Gandhiraj, and M.~Panda, ``Steering {{Angle Prediction}} for
  {{Autonomous Driving}} using {{Federated Learning}}: {{The Impact}} of
  {{Vehicle-To-Everything Communication}},'' in {\em 2021 12th International
  Conference on Computing Communication and Networking Technologies (ICCCNT)},
  pp.~1--7, IEEE, 2021.

\bibitem{pitropov_canadian_2021}
M.~Pitropov, D.~Garcia, J.~Evan, Rebello, M.~Smart, C.~Wang, K.~Czarnecki, and
  S.~Waslander, ``Canadian {{Adverse Driving Conditions}} dataset,'' {\em The
  International Journal of Robotics Research}, vol.~40, no.~4-5, pp.~681--690,
  2021.

\bibitem{yu2020bdd100k}
F.~Yu, H.~Chen, X.~Wang, W.~Xian, Y.~Chen, F.~Liu, V.~Madhavan, and T.~Darrell,
  ``{BDD100k: A diverse driving dataset for heterogeneous multitask
  learning},'' in {\em Proceedings of the IEEE/CVF Conference on Computer
  Vision and Pattern Recognition}, pp.~2636--2645, 2020.

\bibitem{moosavi2017trajectory}
S.~Moosavi, B.~Omidvar-Tehrani, and R.~Ramnath, ``Trajectory annotation by
  discovering driving patterns,'' in {\em Proceedings of the 3rd ACM SIGSPATIAL
  Workshop on Smart Cities and Urban Analytics}, pp.~1--4, 2017.

\bibitem{stallkamp2011german}
J.~Stallkamp, M.~Schlipsing, J.~Salmen, and C.~Igel, ``The german traffic sign
  recognition benchmark: a multi-class classification competition,'' in {\em
  The 2011 International Joint Conference on Neural Networks}, pp.~1453--1460,
  IEEE, 2011.

\bibitem{cordts_cityscapes_2016}
M.~Cordts, M.~Omran, S.~Ramos, T.~Rehfeld, M.~Enzweiler, R.~Benenson,
  U.~Franke, S.~Roth, and B.~Schiele, ``The {{Cityscapes Dataset}} for
  {{Semantic Urban Scene Understanding}},'' in {\em Proceedings of the {{IEEE
  Conference}} on {{Computer Vision}} and {{Pattern Recognition}}},
  pp.~3213--3223, 2016.

\bibitem{alberti_idda_2020-1}
E.~Alberti, A.~Tavera, C.~Masone, and B.~Caputo, ``{{IDDA}}: A large-scale
  multi-domain dataset for autonomous driving,'' {\em IEEE Robotics and
  Automation Letters}, vol.~5, no.~4, pp.~5526--5533, 2020.

\bibitem{kesten2019lyft}
R.~Kesten, M.~Usman, J.~Houston, T.~Pandya, K.~Nadhamuni, A.~Ferreira, M.~Yuan,
  B.~Low, A.~Jain, P.~Ondruska, {\em et~al.}, ``Lyft level 5 av dataset 2019,''
  {\em urlhttps://level5. lyft. com/dataset}, vol.~1, p.~3, 2019.

\bibitem{han_federated_2022}
M.~Han, K.~Xu, S.~Ma, A.~Li, and H.~Jiang, ``Federated learning-based
  trajectory prediction model with privacy preserving for intelligent
  vehicle,'' {\em International Journal of Intelligent Systems}, vol.~37,
  no.~12, pp.~10861--10879, 2022.

\bibitem{dot2018next}
U.~DOT, ``{Next generation simulation (NGSIM) vehicle trajectories and
  supporting data},'' {\em US Department of Transportation}, 2018.

\bibitem{farm_2016}
{State Farm}, ``State farm distracted driver detection,'' {\em Kaggle}, Apr.
  2016.

\bibitem{Naphade22AIC22}
M.~Naphade, S.~Wang, D.~C. Anastasiu, Z.~Tang, M.~Chang, Y.~Yao, L.~Zheng,
  M.~S. Rahman, A.~Venkatachalapathy, A.~Sharma, Q.~Feng, V.~Ablavsky,
  S.~Sclaroff, P.~Chakraborty, A.~Li, S.~Li, and R.~Chellappa, ``The 6th ai
  city challenge,'' in {\em 2022 IEEE/CVF Conference on Computer Vision and
  Pattern Recognition Workshops (CVPRW)}, pp.~3346--3355, IEEE Computer
  Society, June 2022.

\bibitem{martin2019drive}
M.~Martin, A.~Roitberg, M.~Haurilet, M.~Horne, S.~Rei{\ss}, M.~Voit, and
  R.~Stiefelhagen, ``Drive\&act: A multi-modal dataset for fine-grained driver
  behavior recognition in autonomous vehicles,'' in {\em Proceedings of the
  IEEE/CVF International Conference on Computer Vision}, pp.~2801--2810, 2019.

\bibitem{pan2007eyeblink}
G.~Pan, L.~Sun, Z.~Wu, and S.~Lao, ``Eyeblink-based anti-spoofing in face
  recognition from a generic webcamera,'' in {\em 2007 IEEE 11th International
  Conference on Computer Vision}, pp.~1--8, IEEE, 2007.

\bibitem{drutarovsky2014eye}
T.~Drutarovsky and A.~Fogelton, ``Eye blink detection using variance of motion
  vectors.,'' in {\em ECCV Workshops (3)}, pp.~436--448, 2014.

\bibitem{liu2021automatic}
Q.~Liu, Q.~Guo, W.~Wang, Y.~Zhang, and Q.~Kang, ``An automatic detection
  algorithm of metro passenger boarding and alighting based on deep learning
  and optical flow,'' {\em IEEE Trans. Instrum. Meas.}, vol.~70, pp.~1--13,
  Jan. 2021.

\bibitem{Highway_Statistics_2020}
``Table {HM-12}---{Highway Statistics} 2020---{Policy} {Federal Highway
  Administration}.'' https://www.fhwa.dot.gov/policyinformation
  /statistics/2020/hm12.cfm.

\bibitem{J3016_201806}
{On-Road Automated Driving (ORAD) Committee}, ``Taxonomy and definitions for
  terms related to driving automation systems for on-road motor vehicles,''
  June 2018.

\bibitem{gidado_survey_2020}
U.~M. Gidado, H.~Chiroma, N.~Aljojo, S.~Abubakar, S.~I. Popoola, and M.~A.
  {Al-Garadi}, ``A {{Survey}} on {{Deep Learning}} for {{Steering Angle
  Prediction}} in {{Autonomous Vehicles}},'' {\em IEEE Access}, vol.~8,
  pp.~163797--163817, 2020.

\bibitem{bojarski2016end}
M.~Bojarski, D.~Del~Testa, D.~Dworakowski, B.~Firner, B.~Flepp, P.~Goyal, L.~D.
  Jackel, M.~Monfort, U.~Muller, J.~Zhang, {\em et~al.}, ``End to end learning
  for self-driving cars,'' {\em arXiv preprint arXiv:1604.07316}, 2016.

\bibitem{huang_survey_2022}
Y.~Huang, J.~Du, Z.~Yang, Z.~Zhou, L.~Zhang, and H.~Chen, ``A {{Survey}} on
  {{Trajectory-Prediction Methods}} for {{Autonomous Driving}},'' {\em IEEE
  Transactions on Intelligent Vehicles}, vol.~7, pp.~652--674, Sept. 2022.

\bibitem{rjoub_explainable_2022}
G.~Rjoub, J.~Bentahar, and O.~A. Wahab, ``Explainable {{AI-based Federated Deep
  Reinforcement Learning}} for {{Trusted Autonomous Driving}},'' in {\em 2022
  {{International Wireless Communications}} and {{Mobile Computing}}
  ({{IWCMC}})}, pp.~318--323, May 2022.

\bibitem{arnold_survey_2019}
E.~Arnold, O.~Y. Al-Jarrah, M.~Dianati, S.~Fallah, D.~Oxtoby, and
  A.~Mouzakitis, ``A survey on 3d object detection methods for autonomous
  driving applications,'' {\em IEEE Transactions on Intelligent Transportation
  Systems}, vol.~20, no.~10, pp.~3782--3795, 2019.

\bibitem{liu_fedvision_2020}
Y.~Liu, A.~Huang, Y.~Luo, H.~Huang, Y.~Liu, Y.~Chen, L.~Feng, T.~Chen, H.~Yu,
  and Q.~Yang, ``{{FedVision}}: {{An}} online visual object detection platform
  powered by federated learning,'' {\em Proceedings of the AAAI Conference on
  Artificial Intelligence}, vol.~34, pp.~13172--13179, Apr. 2020.

\bibitem{wang2022edge}
S.~Wang, Y.~Hong, R.~Wang, Q.~Hao, Y.-C. Wu, and D.~W.~K. Ng, ``Edge federated
  learning via unit-modulus over-the-air computation,'' {\em IEEE Transactions
  on Communications}, vol.~70, no.~5, pp.~3141--3156, 2022.

\bibitem{wang_federated_2022}
S.~Wang, C.~Li, D.~W.~K. Ng, Y.~C. Eldar, H.~V. Poor, Q.~Hao, and C.~Xu,
  ``Federated {{Deep Learning Meets Autonomous Vehicle Perception}}: {{Design}}
  and {{Verification}},'' {\em IEEE Network}, pp.~1--10, 2022.

\bibitem{perez-gil_deep_2022}
{\'O}.~{P{\'e}rez-Gil}, R.~Barea, E.~{L{\'o}pez-Guill{\'e}n}, L.~M. Bergasa,
  C.~{G{\'o}mez-Hu{\'e}lamo}, R.~Guti{\'e}rrez, and A.~{D{\'i}az-D{\'i}az},
  ``Deep reinforcement learning based control for {{Autonomous Vehicles}} in
  {{CARLA}},'' {\em Multimedia Tools and Applications}, vol.~81,
  pp.~3553--3576, Jan. 2022.

\bibitem{taik_clustered_2022}
A.~Ta{\"i}k, Z.~Mlika, and S.~Cherkaoui, ``Clustered {{Vehicular Federated
  Learning}}: {{Process}} and {{Optimization}},'' {\em IEEE Transactions on
  Intelligent Transportation Systems}, vol.~23, pp.~25371--25383, Dec. 2022.

\bibitem{wang_content-based_2021}
S.~Wang, F.~Liu, and H.~Xia, ``Content-based {{Vehicle Selection}} and
  {{Resource Allocation}} for {{Federated Learning}} in {{IoV}},'' in {\em 2021
  {{IEEE Wireless Communications}} and {{Networking Conference Workshops}}
  ({{WCNCW}})}, ({Nanjing, China}), pp.~1--7, {IEEE}, Mar. 2021.

\bibitem{tianqing2021resource}
Z.~Tianqing, W.~Zhou, D.~Ye, Z.~Cheng, and J.~Li, ``Resource allocation in iot
  edge computing via concurrent federated reinforcement learning,'' {\em IEEE
  Internet of Things Journal}, vol.~9, no.~2, pp.~1414--1426, 2021.

\bibitem{albelaihi_green_2022}
R.~Albelaihi, L.~Yu, W.~D. Craft, X.~Sun, C.~Wang, and R.~Gazda, ``Green
  {{Federated Learning}} via {{Energy-Aware Client Selection}},'' in {\em
  {{GLOBECOM}} 2022 - 2022 {{IEEE Global Communications Conference}}},
  pp.~13--18, Dec. 2022.

\bibitem{nishio_client_2019}
T.~Nishio and R.~Yonetani, ``Client {{Selection}} for {{Federated Learning}}
  with {{Heterogeneous Resources}} in {{Mobile Edge}},'' in {\em {{ICC}} 2019 -
  2019 {{IEEE International Conference}} on {{Communications}} ({{ICC}})},
  pp.~1--7, May 2019.

\end{thebibliography}
\end{document}